\documentclass[10pt, conference, a4paper]{IEEEtran}
\IEEEoverridecommandlockouts
\usepackage{cite}
\usepackage{amsmath,amssymb,amsfonts}
\usepackage{bm}
\usepackage{mathtools}
\usepackage{algorithmic}
\usepackage{graphicx}
\usepackage{textcomp}
\usepackage{xcolor}
\def\BibTeX{{\rm B\kern-.05em{\sc i\kern-.025em b}\kern-.08em
        T\kern-.1667em\lower.7ex\hbox{E}\kern-.125emX}}
\usepackage{lipsum}
\usepackage{tikz}
\usepackage{pgfplots}

\usetikzlibrary{positioning,calc}
\usetikzlibrary{arrows.meta}

\pgfplotsset{compat=1.16}
\pgfplotsset{table/search path={data}}

\usepgfplotslibrary{groupplots}
\usepgfplotslibrary{polar}

\newcommand\colorarea[3][]{%
    \fill[#1]
    ({rel axis cs:0,0}-|{axis cs:#2,0})
    rectangle
    ({rel axis cs:0,1}-|{axis cs:#3,0});
}

\newcommand\addcolorareas[2][]{%
    \pgfplotsforeachungrouped \i/\j in {#2}{
        \edef\temp{\noexpand\colorarea[#1]{\i}{\j}}
        \temp
    }
}

\def\removeleadingzeros#1{\if0#1 \expandafter\else#1\fi}

\def\transformtime#1:#2:#3!{
    \pgfkeys{/pgf/fpu=true,/pgf/fpu/output format=fixed}
    \pgfmathparse{\removeleadingzeros#1*3600-\pgfkeysvalueof{/pgfplots/timeplot 
    zero}*3600+\removeleadingzeros#2*60+\removeleadingzeros#3}
    \pgfkeys{/pgf/fpu=false}
}

\pgfplotsset{
    timeplot zero/.initial=0,
    timeplot/.style={
        x coord trafo/.code={\expandafter\transformtime##1!},
        x coord inv trafo/.code={%
            \pgfkeys{/pgf/fpu=true,/pgf/fpu/output format=fixed}
            \pgfmathsetmacro\hours{floor(##1/3600)+\pgfkeysvalueof{/pgfplots/timeplot
             zero}}
            \pgfmathsetmacro\minutes{floor((##1-(\hours-\pgfkeysvalueof{/pgfplots/timeplot
             zero})*3600)/60)}
            \pgfmathsetmacro\seconds{##1-floor(##1/60)*60}
            \def\pgfmathresult{\pgfmathparse{mod(\hours,60)<10?"0":{},int(mod(\hours,60))}\pgfmathresult:\pgfmathparse{mod(\minutes,60)<10?"0":{},int(mod(\minutes,60))}\pgfmathresult:\pgfmathparse{mod(\seconds,60)<10?"0":{},int(mod(\seconds,60))}\pgfmathresult}
            \pgfkeys{/pgf/fpu=false}
        },
        scaled x ticks=false,
        xticklabel=\tick
    }
}

\newcommand{\mW}{\text{mW}}
\newcommand{\transpose}{^\mathbf{T}}

\definecolor{DLRBlack}{gray}{0}
\definecolor{DLRGrey}{gray}{0.420} 
\colorlet{DLRGray}{DLRGrey}
\definecolor{DLRWhite}{gray}{1}

\colorlet{DLRDarkerGrey}{DLRGrey}
\definecolor{DLRDarkGrey}{gray}{0.537} 
\definecolor{DLRMediumGrey}{gray}{0.702} 
\definecolor{DLRLightGrey}{gray}{0.820} 
\definecolor{DLRLighterGrey}{gray}{0.929} 

\colorlet{DLRDarkerGray}{DLRDarkerGrey}
\colorlet{DLRDarkGray}{DLRDarkGrey}
\colorlet{DLRMediumGray}{DLRMediumGrey}
\colorlet{DLRLightGray}{DLRLightGrey}
\colorlet{DLRLighterGray}{DLRLighterGrey}

\definecolor{DLRDarkerBlue}{RGB}{0, 106, 144}
\definecolor{DLRDarkBlue}{RGB}{0, 156, 208}
\definecolor{DLRBlue}{RGB}{33, 187, 223}
\colorlet{DLRMediumBlue}{DLRBlue}
\definecolor{DLRLightBlue}{RGB}{149, 212, 238}
\definecolor{DLRLighterBlue}{RGB}{201, 232, 251}

\definecolor{DLRDarkerGreen}{RGB}{115, 163, 63}
\definecolor{DLRDarkGreen}{RGB}{158, 193, 76}
\definecolor{DLRGreen}{RGB}{199, 214, 84}
\colorlet{DLRMediumGreen}{DLRGreen}
\definecolor{DLRLightGreen}{RGB}{215, 223, 116}
\definecolor{DLRLighterGreen}{RGB}{228, 234, 173}

\definecolor{DLRDarkerYellow}{RGB}{224, 177, 57}
\definecolor{DLRDarkYellow}{RGB}{254, 206, 73}
\definecolor{DLRYellow}{RGB}{255, 223, 73}
\colorlet{DLRMediumYellow}{DLRYellow}
\definecolor{DLRLightYellow}{RGB}{255, 234, 117}
\definecolor{DLRLighterYellow}{RGB}{255, 248, 189}

\title{%
Learning Antenna Pointing Correction in Operations:
Efficient Calibration of a Black Box
}

\author{
\IEEEauthorblockN{Leif Bergerhoff}
\IEEEauthorblockA{%
    \textit{German Aerospace Center (DLR)}\\
    \textit{German Remote Sensing Data Center (DFD)}\\
    \textit{National Ground Segment (NBS)}\\
    Neustrelitz, Germany\\
    Leif.Bergerhoff@dlr.de}
}

\begin{document}

\maketitle


\begin{abstract}
We propose an efficient offline pointing calibration method for operational 
antenna systems which does not require any downtime.
Our approach minimizes the calibration effort and exploits technical signal 
information which is typically used for monitoring and control purposes in 
ground station operations.
Using a standard antenna interface and data from an operational satellite contact, 
we come up with a robust 
strategy for training data set 
generation. On top of this, we learn the parameters of a suitable coordinate 
transform by means of linear regression.
In our experiments, we show the usefulness of the method in a real-world setup.
\end{abstract}



\section{Introduction}
An accurate calibration of antenna pointing is crucial for robust and reliable 
communications between a ground station and satellites.
To maximize the received signal strength from the satellite, people came up 
with various strategies to optimize the orientation of ground station antennas.
A traditional method is the so-called step track \cite{Dy98} technique.
Extensions and numerous antenna pointing strategies have been suggested (see 
e.g. 
\cite{Ho87,DZLY17,LYT10,MV16,Ba73,MBH68}  and the references therein).
These include applications to stationary \cite{MBH68} as well as mobile systems 
\cite{LYT10}.
Usually, the sun or other cosmic radio sources serve as a reference in order to 
estimate the desired antenna and environmental parameters.
This involves the measurement of offset angles, which, however, often entails an 
interruption of operations and requires human intervention.
Typical parameters include the hardware's mechanical structure, 
weather conditions, and the target position in the sky~\cite{NRPS24}.

At the ground station Neustrelitz, the German Remote Sensing Data Center 
operates multiple antenna systems for payload data acquisition from 
remote-sensing satellites. Satellite communication takes place in different 
frequency bands: L-, S-, X- and Ka-band, i.e. within a frequency range from 1 
GHz to 40 GHz (cf. \cite{IEEE19}).
One of the antennas \cite{Si19} is part of the Ionosphere Monitoring and 
Prediction Center (IMPC) \cite{KB20} and contributes to the Real Time Solar 
Wind (RTSW) observation network \cite{SWPC24}. Within this
context we are engaged in the S-band data reception 
of the ACE~\cite{SFMCMOS98} and DSCOVR~\cite{BB12} satellites.
Both satellites are positioned at the Sun-Earth $L_1$ Lagrange point which 
means that -- from a ground station perspective --
the trajectories roughly correspond to the position of the sun.
As a consequence, our antenna system is used for data reception from sunrise to 
sunset.
From a scientific and practical viewpoint, it is desirable to prevent data 
loss and antenna unavailabilities in order to be informed about solar events and 
their potential impact on infrastructure in space and on earth (see e.g. 
\cite{KB20}).

\subsection{Our Contributions}
We come up with a versatile calibration strategy for semi-automated pointing 
recalibration of antenna systems. 
The primary focus of our approach is the applicability
in parallel to the operational satellite data reception, i.e. without signal loss.
Furthermore, it requires no special knowledge about the construction and the 
implementation of the antenna system. Our method 
treats the device as a black box, i.e. a plain antenna with two 
degrees of freedom and a simple standard interface.
We suggest a reasonable coordinate transform for antenna pointing correction and
describe the parameter learning using a linear regression approach.
In our experiments, we show the applicability to an antenna system in a 
real-world scenario and 
prove the usefulness and optimality of our approach.
To the best of our knowledge, this is the most comprehensive method considering
antenna parameter learning, technical signal information processing, and 
operational satellite data reception at the same time.

\subsection{Structure of the Paper}
In Section~\ref{sec:tech} we discuss the basic properties of 
our black box antenna system and the employed technical signal information.
Section~\ref{sec:theory} introduces the required coordinate transform and  
linear regression model.

In Section~\ref{sec:exp}, we develop a calibration strategy and prove their 
usefulness in experiments.

We conclude with a summary and outlook in Section~\ref{sec:sum}.


\section{Technical Background}
\label{sec:tech}
\begin{figure}
\centering
\begin{tikzpicture}
\begin{groupplot}[%
    group style={group size=1 by 2, horizontal sep=0mm, vertical sep=14mm},
    width=\columnwidth,
    height=.4\columnwidth,
    xlabel={UTC [HH:MM:SS]},
    xtick={{07:00:00},{09:00:00},{11:00:00.5},{13:00:00.5},{15:00:00},{17:00:00.5}},
    /tikz/font=\tiny,
    grid=both,
    minor tick num=1,
    timeplot,
    timeplot zero=0]
\nextgroupplot[%
ylabel={Azimuth [$^\circ$]}]
\addplot+[DLRDarkerBlue, mark=none, thick]
table [x index=0, y index=1]
{dscovr_24044_pointing.txt};
\nextgroupplot[%
ylabel={Elevation [$^\circ$]}]
\addplot+[DLRDarkerBlue, mark=none, thick]
table [x index=0, y index=2]
{dscovr_24044_pointing.txt};
\end{groupplot}
\end{tikzpicture}
\caption{Exemplary antenna pointing towards the DSCOVR satellite.}
\label{fig:dscovr_track}
\end{figure}
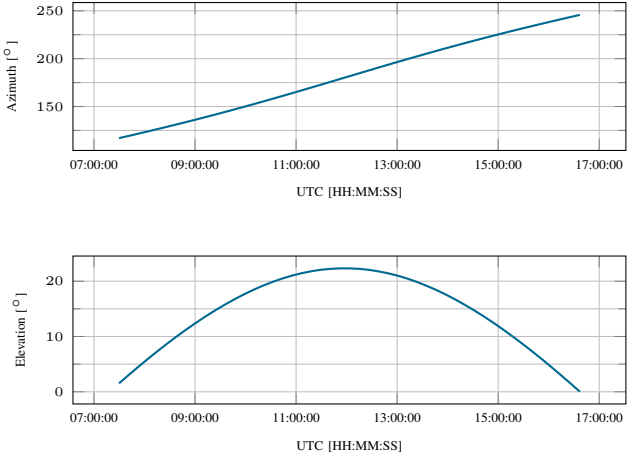

We aim for a software-based calibration procedure which does not need any 
special knowledge about the implementation of the antenna system.
For this purpose, we consider an antenna with two degrees of freedom 
(azimuth and elevation axes) and a simple interface for tracking information. 
It permits 
to upload a tracking table which 
describes the antenna orientation over time. In our case, this corresponds to 
the trajectory of the DSCOVR satellite from the antenna's point of view.
Table~\ref{tab:track_tab} shows the content of a typical tracking table, 
containing columns for \textit{time}, \textit{azimuth}, and \textit{elevation} 
angle.
\begin{table}
\renewcommand{\arraystretch}{1.3}
\caption{Excerpt from a typical tracking table}
\label{tab:track_tab}
\centering
\begin{tabular}{lrr}
\hline
Time (UTC) & Azimuth [$^\circ$] & Elevation [$^\circ$]\\
\hline
07:18:21 & 114.67 & 0.00\\
07:29:45 & 116.97 & 1.53\\
07:41:09 & 119.28 & 3.03\\
\hline
\end{tabular}
\end{table}
A standard antenna system aligns with the specified tracking points and 
interpolates the pointing directions in between.
Our antenna supports the specification of up to 100 different track points within 
one file.
Fig.~\ref{fig:dscovr_track} visualizes an exemplary antenna pointing for DSCOVR 
data reception over time.

\begin{figure}
\centering
\begin{tikzpicture}
\node[draw, rectangle, black, minimum height=16mm] (box0) {%
\parbox{.25\columnwidth}{%
\centering
\scriptsize
File-based
Pointing Information\\
\textit{(Tracking Table)}
}
};
\node[draw, rectangle, DLRMediumGrey, fill=DLRLighterGrey, text=black, minimum 
height=16mm, dashed, line width=1pt, right=of box0] 
(box1) {
\parbox{.25\columnwidth}{%
\centering
\scriptsize
File Upload to Antenna System
}
};
\node[draw, rectangle, DLRMediumGrey, fill=DLRLighterGrey, text=black, minimum 
height=16mm, dashed, line width=1pt, below=10mm of box1] (box2) {
\parbox{.25\columnwidth}{%
\centering
\scriptsize
Physical\\
Antenna Motion
}
};
\node[draw, rectangle, black, minimum height=16mm, left=of box2] (box3) {
\parbox{.25\columnwidth}{%
\centering
\scriptsize
Quality Information Monitoring System\\
\textit{(Antenna Pointing, Signal Level)}
}
};
\draw[-{Classical TikZ Rightarrow[scale=2]}, line width=1pt, DLRMediumGrey] 
(box0) -- (box1);
\draw[-{Classical TikZ Rightarrow[scale=2]}, line width=1pt, DLRMediumGrey] 
(box1) -- (box2);
\draw[-{Classical TikZ Rightarrow[scale=2]}, line width=1pt, DLRMediumGrey] 
(box2) -- (box3);
\draw[-{Classical TikZ Rightarrow[scale=2]}, line width=1pt, black, dashed] 
(box0) -- (box3);
\end{tikzpicture}
\caption{The basic evaluation process for antenna pointing quality.}
\label{fig:basic_evaluation}
\end{figure}
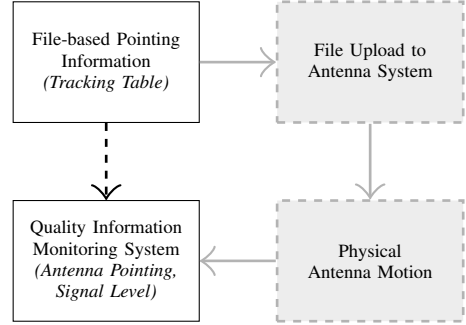

During operations, the related ground station hardware and 
software supplies technical signal information. Traditionally, ground station 
processes use this data for monitoring and control purposes. 
We store this knowledge and employ a small subset of the data, namely 
the signal level information, as a measure for the reception quality.
Our signal levels are given in dBm (decibel-milliwatts). Given a power P in mW, 
the corresponding power level x in dBm is calculated as 
follows:
\begin{equation} \label{eq:dbm}
    x = 10 \cdot \log_{10} \frac{P}{1~\mW} \,.
\end{equation}
In combination with the previously mentioned tracking table, this allows us to 
draw conclusions about the influence of the antenna pointing on the measured 
satellite signal strength.
In Fig.~\ref{fig:basic_evaluation}, we illustrate our signal quality assessment 
process: We use a static tracking table which we upload to the antenna system in 
advance of the satellite data reception. As mentioned before, the antenna points 
to the specified directions during operations and interpolates in between.
Note, that using this simple -- file-based -- standard interface it is not 
possible 
to change 
the antenna orientation dynamically.
This would require the upload of 
new pointing information to the system and interrupts the operational 
satellite 
communication.
In parallel to the satellite data reception, we measure and store the received 
signal level over time.


\section{Theory and Modeling}
\label{sec:theory}
The main task of our antenna pointing calibration method is the estimation of a 
suitable 
coordinate transform
\begin{equation} \label{eq:trans}
\bm{y} := f(\bm{x}) := \mathbf{T} \bm{x}
\end{equation}
for optimal antenna orientation.
The coordinate transform $f$ allows to estimate $p$-dimensional device specific 
pointing 
coordinates $\bm{y} \in \mathbb{R}^p$ from the desired pointing directions
$\bm{x} \in \mathbb{R}^p$.
Within this paper, we implement $f$ in terms of a matrix-vector multiplication 
with a real-valued $p \times p$ transformation matrix $\mathbf{T}$.
Furthermore, we make use of homogeneous coordinates (see e.g. \cite{HZ04}) 
in order of being able to describe occurring nonlinear projective geometry in 
terms of linear mappings. 
As an example, we have $p = 3$ and
$\bm{x}=s\,\cdot\,(x^*_1, x^*_2, 1)\transpose$ with arbitrary scalar 
$s \neq 0$ for two-dimensional pointing information $(x^*_1, x^*_2)\transpose$.
For simplicity, we set $s$ to 1. 
The same transfers to $\bm{y}$.

In order to train an appropriate coordinate transform $f$, we consider a multiple 
linear regression model with multiple outputs (cf. \cite{HTF09}):
\begin{equation}
y_k := f_k(\bm{x}) := \sum_{j=1}^{p} t_{k,j} \, x_{j}, 
\quad \forall k \in \{1,\ldots,p\},
\end{equation}
where we consider no bias term since our data is centered.

In order to learn the entries $t_{k,j}$ of $\mathbf{T}$, we do least squares 
approximation, i.e. we minimize the residual sum of squares
\begin{align} \label{eq:rss}
    \text{RSS}(\mathbf{T})
    &= \sum_{i=1}^{N} \sum_{k=1}^{p} (\bm{y}_{i,k} - f_k(\bm{x}_i))^2
\end{align}
using $N$ pairs $(\bm{x}_i, \bm{y}_i)$, i.e. $i = \{1,\ldots,N\}$, as training 
data.

Subsequently, we restrict ourselves to the scenario $p=3$,
making use of azimuth 
$\varphi$ and elevation $\theta$ angles as x- and y-coordinates respectively.
Accordingly, we write the coordinate transform \eqref{eq:trans} as
\begin{equation}
\underbrace{%
    \begin{pmatrix}
        \tilde{\varphi}\\
        \tilde{\theta}\\
        1
\end{pmatrix}}_{:= \bm{y}} = 
\underbrace{%
    \begin{pmatrix}
        t_{1,1} & t_{1,2} & t_{1,3}\\
        t_{2,1} & t_{2,2} & t_{2,3}\\
        t_{3,1} & t_{3,2} & t_{3,3}\\
\end{pmatrix}}_{:= \mathbf{T}}
\underbrace{%
    \begin{pmatrix}
        \varphi\\
        \theta\\
        1
\end{pmatrix}}_{:= \bm{x}}.
\end{equation}

Thinking in terms of a sphere surface -- with arbitrary radius larger than zero 
-- we have  $\varphi \in [0, 2 \pi)$ and $\theta \in [-\pi, \pi]$. 
In terms of cardinal directions, we have
$\varphi = 0$ (north),
$\varphi = \frac{\pi}{2}$ (east),
$\varphi = \pi$ (south), and 
$\varphi = \frac{3}{2}\pi$ (west).
On the other hand, the $\theta = 0$ plane refers to the earth 
surface and $\theta = \pi$ refers to a pointing straight into the sky.
We support negative elevation angles for correction purposes. Typically, 
antenna systems also support small negative elevation values.


\section{Experiments and Calibration Strategy}
\label{sec:exp}
\subsection{Step Track}
\subsubsection{Training}
In our first experiment, we aim for the estimation of a transformation matrix 
$\mathbf{T}$ by following a manual step track \cite{Dy98} procedure: We use the 
electromagnetic radiation of the sun in order to estimate locally optimal 
azimuth and elevation angles. Meanwhile, the antenna operates 
in \textit{sun track mode} such that it aligns with the sun (e.g. 
using track information based on \cite{PFWB21}).
In our setup, it takes approximately 4 minutes to estimate a locally optimal 
pointing direction $\bm{y}$ for a given pointing $\bm{x}$.
Throughout one day, we estimate $N=60$ pairs $(\bm{x},\bm{y})$ which serve as 
our training data set.
Accordingly, we use the latter to minimize \eqref{eq:rss} and learn a matrix 
$\mathbf{T}$.
The corresponding training error is given in Table~\ref{tab:train_err}.
\begin{table}
\renewcommand{\arraystretch}{1.3}
\caption{Training error}
\label{tab:train_err}
\centering
\begin{tabular}{lrr}
\hline
& MAE [$^\circ$] & MSE [$^\circ$]\\
\hline
Azimuth $\varphi$ \textit{(step track)}             & 0.023918 & 0.000772\\
Elevation $\theta$ \textit{(step track)}            & 0.021072 & 0.000662\\
Azimuth $\varphi$ \textit{(improved calibration)}   & 0.073698 & 0.010432\\
Elevation $\theta$ \textit{(improved calibration)}  & 0.064724 & 0.011275\\
\hline
\end{tabular}
\end{table}
The transformation matrix $\mathbf{T}$ reads
\begin{equation}
    \mathbf{T} \approx
    \begin{psmallmatrix*}[r]
        0.994773 & -0.017231 &  0.022903\\[.25ex]
        0.007398 &  0.992050 & -0.016989\\[.25ex]
        0.000000 &  0.000000 &  1.000000\\
    \end{psmallmatrix*}
\end{equation}
and resembles an affine transform.
The key impact factor is a counter-clockwise rotation by 
$\sim0.43^\circ$ in the azimuth-elevation plane.

\subsubsection{Testing}
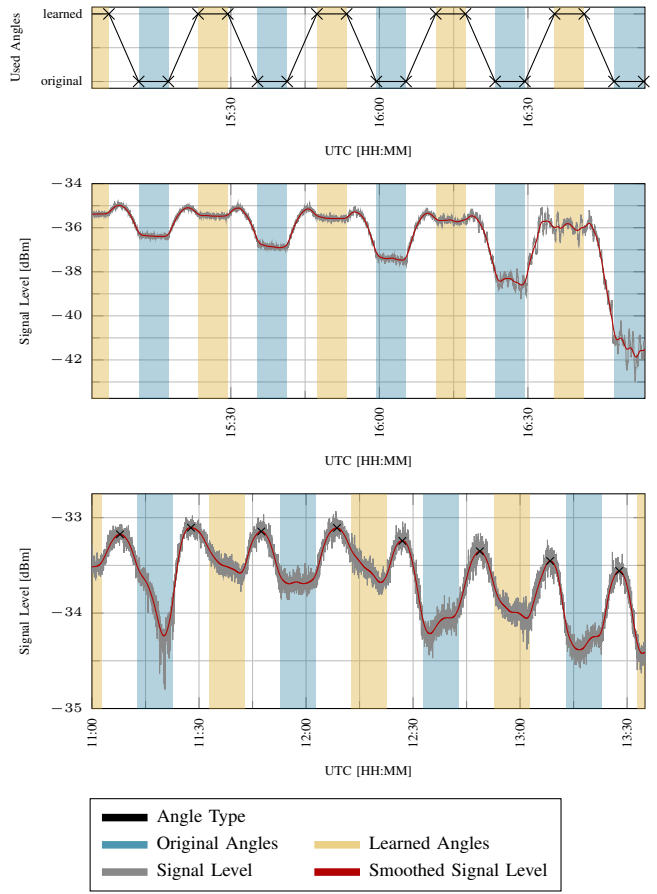
\begin{figure}
\centering
\begin{tikzpicture}
\begin{groupplot}[%
    group style={group size=1 by 3,
        horizontal sep=0mm, vertical sep=8ex},
    width=\columnwidth,
    height=.5\columnwidth,
    scaled x ticks=false,
    x tick label style={rotate=90},
    xlabel={UTC [HH:MM]},
    ylabel={Signal Level [dBm]},
    grid=both,
    minor tick num=1,
    /tikz/font=\tiny]
\nextgroupplot[%
    height=.3\columnwidth,
    xmin=33300,
    xmax=40000,
    restrict x to domain=32700:40000,
    scaled x ticks=false,
    xtick={34982,36782,38582},
    xticklabels={{15:30},{16:00},{16:30}}, 
    ylabel={Used Angles},
    ytick={0,1},
    yticklabels={{original},{learned}}]
\colorarea[DLRDarkerYellow, opacity=.4]{25945}{26305}
\colorarea[DLRDarkerYellow, opacity=.4]{27385}{27745}
\colorarea[DLRDarkerYellow, opacity=.4]{28825}{29185}
\colorarea[DLRDarkerYellow, opacity=.4]{30265}{30625}
\colorarea[DLRDarkerYellow, opacity=.4]{31705}{32065}
\colorarea[DLRDarkerYellow, opacity=.4]{33145}{33505}
\colorarea[DLRDarkerYellow, opacity=.4]{34585}{34945}
\colorarea[DLRDarkerYellow, opacity=.4]{36025}{36385}
\colorarea[DLRDarkerYellow, opacity=.4]{37465}{37825}
\colorarea[DLRDarkerYellow, opacity=.4]{38905}{39265}
\colorarea[DLRDarkerYellow, opacity=.4]{40345}{40705}
\colorarea[DLRDarkerBlue, opacity=.3]{24302}{25585}
\colorarea[DLRDarkerBlue, opacity=.3]{26665}{27025}
\colorarea[DLRDarkerBlue, opacity=.3]{28105}{28465}
\colorarea[DLRDarkerBlue, opacity=.3]{29545}{29905}
\colorarea[DLRDarkerBlue, opacity=.3]{30985}{31345}
\colorarea[DLRDarkerBlue, opacity=.3]{32425}{32785}
\colorarea[DLRDarkerBlue, opacity=.3]{33865}{34225}
\colorarea[DLRDarkerBlue, opacity=.3]{35305}{35665}
\colorarea[DLRDarkerBlue, opacity=.3]{36745}{37105}
\colorarea[DLRDarkerBlue, opacity=.3]{38185}{38545}
\colorarea[DLRDarkerBlue, opacity=.3]{39625}{39985}
\addplot+[DLRBlack, mark=x, mark options={scale=1.5}] table [x=time, y=learned] {
    time  learned
    24302 0
    25585 0
    25945 1
    26305 1
    26665 0
    27025 0
    27385 1
    27745 1
    28105 0
    28465 0
    28825 1
    29185 1
    29545 0
    29905 0
    30265 1
    30625 1
    30985 0
    31345 0
    31705 1
    32065 1
    32425 0
    32785 0
    33145 1
    33505 1
    33865 0
    34225 0
    34585 1
    34945 1
    35305 0
    35665 0
    36025 1
    36385 1
    36745 0
    37105 0
    37465 1
    37825 1
    38185 0
    38545 0
    38905 1
    39265 1
    39625 0
    39985 0
    40345 1
    40705 1
};
\nextgroupplot[%
    xmin=33300,
    xmax=40000,
    restrict x to domain=33300:40000,
    scaled x ticks=false,
    xtick={34982,36782,38582},
    xticklabels={{15:30},{16:00},{16:30}}, 
    legend columns=2,
    legend style={/tikz/every even column/.append style={column 
            sep=1em},
        font=\scriptsize},
    legend image post style={line width=1mm},
    legend cell align={left},
    legend to name=step_track_legend]
%
%
\colorarea[DLRDarkerYellow, opacity=.4]{25945}{26305}
\colorarea[DLRDarkerYellow, opacity=.4]{27385}{27745}
\colorarea[DLRDarkerYellow, opacity=.4]{28825}{29185}
\colorarea[DLRDarkerYellow, opacity=.4]{30265}{30625}
\colorarea[DLRDarkerYellow, opacity=.4]{31705}{32065}
\colorarea[DLRDarkerYellow, opacity=.4]{33145}{33505}
\colorarea[DLRDarkerYellow, opacity=.4]{34585}{34945}
\colorarea[DLRDarkerYellow, opacity=.4]{36025}{36385}
\colorarea[DLRDarkerYellow, opacity=.4]{37465}{37825}
\colorarea[DLRDarkerYellow, opacity=.4]{38905}{39265}
\colorarea[DLRDarkerYellow, opacity=.4]{40345}{40705}
\colorarea[DLRDarkerBlue, opacity=.3]{24302}{25585}
\colorarea[DLRDarkerBlue, opacity=.3]{26665}{27025}
\colorarea[DLRDarkerBlue, opacity=.3]{28105}{28465}
\colorarea[DLRDarkerBlue, opacity=.3]{29545}{29905}
\colorarea[DLRDarkerBlue, opacity=.3]{30985}{31345}
\colorarea[DLRDarkerBlue, opacity=.3]{32425}{32785}
\colorarea[DLRDarkerBlue, opacity=.3]{33865}{34225}
\colorarea[DLRDarkerBlue, opacity=.3]{35305}{35665}
\colorarea[DLRDarkerBlue, opacity=.3]{36745}{37105}
\colorarea[DLRDarkerBlue, opacity=.3]{38185}{38545}
\colorarea[DLRDarkerBlue, opacity=.3]{39625}{39985}
\addlegendimage{no markers, DLRBlack, ultra thick}
\addlegendentry{Angle Type}
\addlegendimage{no markers, DLRWhite, ultra thick}
\addlegendentry{}
\addlegendimage{no markers, DLRDarkerBlue!70, ultra thick}
\addlegendentry{Original Angles}
\addlegendimage{no markers, DLRDarkerYellow!60, ultra thick}
\addlegendentry{Learned Angles}
\addlegendimage{no markers, DLRDarkGrey, ultra thick}
\addlegendentry{Signal Level}
\addlegendimage{no markers, red!70!black, ultra thick}
\addlegendentry{Smoothed Signal Level}
\addplot+[DLRDarkGrey, mark=none] table [x={time}, y={if_lvl}]
{dscovr_20281_136090.small.tab};
\addplot+[red!70!black, mark=none] table [x={time}, y={if_lvl_smth_30}]
{dscovr_20281_136090_smth.small.tab};
\nextgroupplot[%
    restrict x to domain=39600:48900,
    scaled x ticks=false,
    xtick={0, 1800, 3600, 5400, 7200, 9000, 10800, 12600, 
        14400, 16200, 18000, 19800, 21600, 23400, 25200, 27000, 
        28800, 30600, 32400, 34200, 36000, 37800, 39600, 41400, 
        43200, 45000, 46800, 48600, 50400, 52200, 54000, 55800, 
        57600, 59400, 61200, 63000, 64800, 66600, 68400, 70200, 
        72000, 73800, 75600, 77400, 79200, 81000, 82800, 84600},
    xticklabels={{00:00}, {00:30}, {01:00}, {01:30}, {02:00}, 
        {02:30}, {03:00}, {03:30}, {04:00}, {04:30}, {05:00}, 
        {05:30}, {06:00}, {06:30}, {07:00}, {07:30}, {08:00}, 
        {08:30}, {09:00}, {09:30}, {10:00}, {10:30}, {11:00}, 
        {11:30}, {12:00}, {12:30}, {13:00}, {13:30}, {14:00}, 
        {14:30}, {15:00}, {15:30}, {16:00}, {16:30}, {17:00}, 
        {17:30}, {18:00}, {18:30}, {19:00}, {19:30}, {20:00}, 
        {20:30}, {21:00}, {21:30}, {22:00}, {22:30}, {23:00}, 
        {23:30}},
    xmin=39600,
    xmax=48900,
    ymin=-35,
    ymax=-32.75]
%
\colorarea[DLRDarkerYellow, opacity=.4]{39167}{39767}
\colorarea[DLRDarkerYellow, opacity=.4]{41567}{42167}
\colorarea[DLRDarkerYellow, opacity=.4]{43967}{44567}
\colorarea[DLRDarkerYellow, opacity=.4]{46367}{46967}
\colorarea[DLRDarkerYellow, opacity=.4]{48767}{49367}
\colorarea[DLRDarkerYellow, opacity=.4]{51167}{51767}
%
\colorarea[DLRDarkerBlue, opacity=.3]{37967}{38567}
\colorarea[DLRDarkerBlue, opacity=.3]{40367}{40967}
\colorarea[DLRDarkerBlue, opacity=.3]{42767}{43367}
\colorarea[DLRDarkerBlue, opacity=.3]{45167}{45767}
\colorarea[DLRDarkerBlue, opacity=.3]{47567}{48167}
\colorarea[DLRDarkerBlue, opacity=.3]{49967}{50567}
\colorarea[DLRDarkerBlue, opacity=.3]{52367}{52967}
\addplot+[DLRDarkGrey, mark=none] table [x={time}, y={if_lvl}]
{dscovr_2317202_excerpt.tab};
\addplot+[red!70!black, mark=none] table [x={time}, y={if_lvl_s1}]
{dscovr_2317202_excerpt.tab};
\addplot+[DLRBlack, only marks, mark=x, mark options={scale=1}]
table [x={time}, y={if_lvl_s1}]
{dscovr_2317202_extrema_excerpt.tab};
\end{groupplot}
\end{tikzpicture}
\vskip 1ex
\ref{step_track_legend}
\caption{Pointing direction type and measured signal level. \textbf{Top:}~ Used 
angle type over time for step track testing. \textbf{Middle:}~ Step track test 
setup. 
\textbf{Bottom:}~ Reevaluation of the learned matrix $\mathbf{T}$ and detection 
of maxima positions.}
\label{fig:step_track_test}
\end{figure}

In order to evaluate the quality of the estimated pointing correction, we use 
the learned coordinate transform to create an adapted tracking table for 
operational DSCOVR satellite data reception
(see Fig.~\ref{fig:apply_correction}).
As a consequence of the previously mentioned limit of 100 track points, our 
experiment's tracking table consists of 6 minute intervals. 
Therein, we either use the 
originally provided pointing directions, the adapted \textit{(learned)} 
pointing directions, or a convex combination of both \textit{(transition 
intervals)}. A visualization of this selection strategy can be found in the top 
plot of Fig.~\ref{fig:step_track_test}.

We illustrate an excerpt of the resulting signal levels in the middle plot of
Fig.~\ref{fig:step_track_test}. One can see that there is a gain in signal 
level of up to 5~dBm (in the afternoon and for low elevation angles).
According to \eqref{eq:dbm}, this means more than a tripling of the measured 
signal level.
Apart from that, one observes that antenna pointing angles in the transition 
intervals \textit{(white regions)} lead to slightly higher signal levels in 
comparison to the learned directions \textit{(yellow regions)}. This indicates 
that our coordinate 
transformation is not yet 
optimal.

Several months later, we repeat the test in order to reevaluate our coordinate 
transform w.r.t. potentially changed antenna parameters. 
Due to a longer lasting track and the restriction to 100 points in one tracking 
table, we have to increase the interval duration to 10 minutes.
The bottom plot of Fig.~\ref{fig:step_track_test} shows that our previous 
calibration does not lead to significant improvements anymore. Now, better 
pointing directions clearly lie in the transition intervals.
Note that strong signal level changes within the intervals of the original and 
learned angles (e.g. around 11:20 UTC) can be traced back to obstacles in the 
signal line of sight.


\subsection{Improved Calibration Strategy}

\begin{figure}
\centering
\begin{tikzpicture}
\node[draw, rectangle, black, minimum height=16mm] (box0) {%
    \parbox{.22\columnwidth}{%
        \centering
        \scriptsize
        \textbf{Original}\\
        File-based
        Pointing Information\\
        \textit{(Tracking Table)}
    }
};
\node[draw, rectangle, DLRDarkerGreen, fill=DLRDarkGreen, text=black, 
minimum height=16mm, right=of box0] 
(box1) {
    \parbox{.28\columnwidth}{%
        \centering
        \scriptsize
        \textbf{Adapted}\\
        File-based
        Pointing Information\\
        \textit{(Altered Tracking Table)}}
};
\node[draw, rectangle, black, minimum height=16mm, right=of box1] 
(box2) {
    \parbox{.19\columnwidth}{%
        \centering
        \scriptsize
        File Upload to Antenna System
    }
};
\draw[-{Classical TikZ Rightarrow[scale=2]}, line width=1pt, black]
(box0) -- (box1);
\draw[-{Classical TikZ Rightarrow[scale=2]}, line width=1pt, black]
(box1) -- (box2);
\end{tikzpicture}
\caption{Process of using our software-based antenna pointing correction.}
\label{fig:apply_correction}
\end{figure}
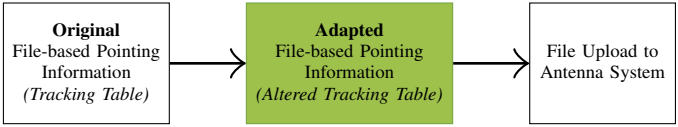

In order to cope with this situation and to prepare for future recalibrations, 
we come up with a semi-automated calibration strategy.
Our procedure is based on the \textit{signal level measurements} 
which we gain during operations as well as on the \textit{original} and 
\textit{adapted pointing information} (cf. Fig.~\ref{fig:apply_correction})).
As before, the tracking table contains alternating intervals with original and  
adapted antenna coordinates.
We suggest the subsequent routine for training data set generation:
\begin{enumerate}
\item Noise reduction: application of a low-pass filter (e.g. Gaussian filter) to 
the signal level measurements
\item Estimation of the signal level maxima
\begin{enumerate}
\item Detection of preliminary signal level maxima (e.g. based on negative second 
derivatives)
\item Merging of related preliminary maxima (e.g. using mean shift clustering 
\cite{Ch95} w.r.t. time)
\item Optimization of the estimated cluster positions using a first-order 
optimization method like the heavy ball method \cite{Po87}
\item Merging of related optimized maxima positions (e.g. using mean shift 
clustering)
\end{enumerate}
\end{enumerate}
Next, the corresponding intended and actual pointing coordinates at 
the times of the detected signal level maxima can be used to learn the matrix 
$\mathbf{T}$ (i.e. the minimization of \eqref{eq:rss}).


\subsection{Improved Calibration}
\subsubsection{Training}

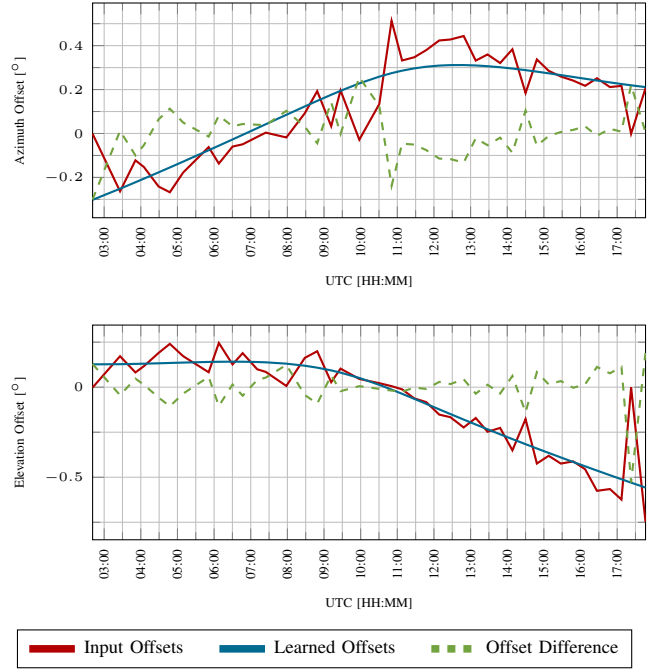
\begin{figure}
\centering
\begin{tikzpicture}
\begin{groupplot}[%
    group style={group size=1 by 2,
        horizontal sep=24mm, vertical sep=9ex},
    width=\columnwidth,
    height=.5\columnwidth,
    restrict x to domain=9665:63987,
    xmin=9665,
    xmax=63987,
    scaled x ticks=false,
    xtick={0, 3600, 7200, 10800, 
        14400, 18000, 21600, 25200, 
        28800, 32400, 36000, 39600, 
        43200, 46800, 50400, 54000, 
        57600, 61200, 64800, 68400, 
        72000, 75600, 79200, 82800},
    xticklabels={{00:00}, {01:00}, {02:00}, 
        {03:00}, {04:00}, {05:00}, 
        {06:00}, {07:00}, {08:00}, 
        {09:00}, {10:00}, {11:00}, 
        {12:00}, {13:00}, {14:00}, 
        {15:00}, {16:00}, {17:00}, 
        {18:00}, {19:00}, {20:00}, 
        {21:00}, {22:00}, {23:00}},
    x tick label style={rotate=90},
    xlabel={UTC [HH:MM]},
    /tikz/font=\tiny]
\nextgroupplot[%
    ylabel={Azimuth Offset [$^\circ$]},
    grid=both,
    minor tick num=1,
    legend columns=3,
    legend style={/tikz/every even column/.append style={column 
            sep=1em},
        font=\scriptsize},
    legend image post style={line width=1mm},
    legend cell align={left},
    legend to name=offset_improved_legend]
\addlegendimage{no markers, red!70!black, ultra thick}
\addlegendentry{Input Offsets}
\addlegendimage{no markers, DLRDarkerBlue, ultra thick}
\addlegendentry{Learned Offsets}
\addlegendimage{no markers, DLRDarkerGreen, ultra thick, dashed}
\addlegendentry{Offset Difference}
\addplot+[red!70!black, mark=none, thick] table [x={time}, y={daz_in_if}]
{dscovr_2317202_angles_extrema.tab};
\addplot+[DLRDarkerBlue, mark=none, thick] table [x={time}, y={daz_in_pred}]
{dscovr_2317202_angles_extrema.tab};
\addplot+[DLRDarkerGreen, thick, dashed, mark=none] table [x={time}, 
y={daz_if_pred}]
{dscovr_2317202_angles_extrema.tab};    
\nextgroupplot[%
    ylabel={Elevation Offset [$^\circ$]},
    grid=both,
    minor tick num=1]
\addplot+[red!70!black, mark=none, thick] table [x={time}, y={del_in_if}]
{dscovr_2317202_angles_extrema.tab};
\addplot+[DLRDarkerBlue, mark=none, thick] table [x={time}, y={del_in_pred}]
{dscovr_2317202_angles_extrema.tab};
\addplot+[DLRDarkerGreen, thick, dashed, mark=none] table [x={time}, 
y={del_if_pred}]
{dscovr_2317202_angles_extrema.tab}; 
\end{groupplot}
\end{tikzpicture}
\vskip 1ex
\ref{offset_improved_legend}
\caption{Estimated and learned offsets using the improved calibration strategy. 
\textbf{Top:}~ Azimuth angle offsets. \textbf{Bottom:}~ Elevation angle 
offsets.}
\label{fig:23_doy_172_offsets}
\end{figure}

Now, we can directly apply the suggested calibration strategy to the 
reevaluation test data (see bottom plot in Fig.~\ref{fig:step_track_test}).
We are able to detect $N=42$ signal level maxima (illustrated as 
crosses). Accordingly, we use the related original and adapted pointing 
information to train the new transformation matrix
\begin{equation}
\mathbf{T} \approx
\begin{psmallmatrix*}[r]
0.997936 & -0.005520 &  0.007442\\[.25ex]
0.002914 &  0.995512 & -0.005053\\[.25ex]
0.000000 &  0.000000 &  1.000000
\end{psmallmatrix*}\,.
\end{equation}
The corresponding training error is also listed in Tab.~\ref{tab:train_err}.
It is slightly higher than for our manual step track calibration, 
for reasons visible in
Fig.~\ref{fig:23_doy_172_offsets}.
Therein, we depict the estimated as well as the
learned azimuth and elevation offsets throughout the satellite data reception.
One can observe two outliers with significant impact on the error values:
in the azimuth offset at around 10:50~UTC and in
the elevation offset at around 17:23~UTC.
While the first outlier results from a signal level change induced by the 
satellite, the second can be traced back to low elevation values and obstacles 
close to the ground.
Overall, the learning has a denoising and stabilizing effect on 
the offset estimation.
This is desirable and limits the impact of the outliers.
Again, the matrix $\mathbf{T}$ implements an affine transform in the 
azimuth-elevation plane, i.e.
\begin{equation}
\begin{psmallmatrix*}[r]
\tilde{\varphi}\\[.25ex]
\tilde{\theta}
\end{psmallmatrix*}
=
\underbrace{%
\begin{psmallmatrix*}[r]
0.997936 & -0.005520\\[.25ex]  
0.002914 &  0.995512\\
\end{psmallmatrix*}}_{=: \mathbf{A}}
\begin{psmallmatrix*}[r]
\varphi\\[.25ex]
\theta
\end{psmallmatrix*}
+
\underbrace{%
\begin{psmallmatrix*}[r]
 0.007442\\[.25ex]
-0.005053\\
\end{psmallmatrix*}}_{=: \bm{t}} \,,
\end{equation}
where $\bm{t}$ denotes a translation. The matrix $\mathbf{A}$ describes -- in 
this order --
a scaling $\mathbf{S_1}$, a shearing 
$\mathbf{S_2}$, and a counter-clockwise rotation $\mathbf{R}$
\begin{align}
\mathbf{S_1} & =
\begin{psmallmatrix*}[r]
0.997940 & 0.000000\\[.25ex]
0.000000 & 0.995524\\ 
\end{psmallmatrix*}\\[.5ex]
\mathbf{S_2} & =
\begin{psmallmatrix*}[r]
1.000000 & -0.002625\\[.25ex]
0.000000 &  1.000000\\ 
\end{psmallmatrix*}\\[.5ex]
\mathbf{R} & =
\begin{psmallmatrix*}[r]
0.999996 & -0.002920\\[.25ex]
0.002920 &  0.999996\\
\end{psmallmatrix*}
\end{align}
in the azimuth elevation plane, s.t. $\mathbf{A} := \mathbf{R} \mathbf{S_2} 
\mathbf{S_1}$.
This time, the key impact factor is a counter-clockwise rotation by 
$0.167279^\circ$.

\subsubsection{Testing}
\begin{figure}
    \centering
    \begin{tikzpicture}
        \begin{polaraxis}[%
            width=.6\columnwidth,
            height=.6\columnwidth,
            xtick distance=45,
            ytick={1},
            yticklabels=\empty,
            /tikz/font=\scriptsize]
            \addplot[%
            DLRBlack,
            domain=0:45,
            fill=DLRYellow,
            fill opacity=.5]
            {0.5}--(0,0)--cycle;
            \addplot+[DLRDarkerBlue, mark=x, mark size=4, every mark/.append 
            style={thick}] 
            coordinates {
                (0,0) (0,1)
                (0,0) (45,1)
                (0,0) (90,1)
                (0,0) (135,1)
                (0,0) (180,1)
                (0,0) (225,1)
                (0,0) (270,1)
                (0,0) (315,1)
                (0,0)};
            \node[DLRBlack] (angle) at (axis cs:22.5,0.33) {\large 
            $\bm{\alpha}$};
            \addplot+[DLRDarkerBlue, mark=none, ultra thick] 
            coordinates {
                (0,0) (0,1)
            };
            \addplot+[DLRDarkerGray, mark=none] 
            coordinates {
                (90,{sin(45)}) (45,1)
            };
            \addplot+[red, mark=none] 
            coordinates {
                (45,1) (0,{cos(45)})
            };
            \node[DLRDarkerBlue] (1) at (axis cs:10,1) {\textbf{1}};
            \node[DLRDarkerBlue] (3) at (axis cs:55,1) {\textbf{3}};
            \node[DLRDarkerBlue] (5) at (axis cs:100,1) {\textbf{5}};
            \node[DLRDarkerBlue] (7) at (axis cs:145,1) {\textbf{7}};
            \node[DLRDarkerBlue] (9) at (axis cs:190,1) {\textbf{9}};
            \node[DLRDarkerBlue] (11) at (axis cs:235,1) {\textbf{11}};
            \node[DLRDarkerBlue] (13) at (axis cs:280,1) {\textbf{13}};
            \node[DLRDarkerBlue] (15) at (axis cs:325,1) {\textbf{15}};
            \node[DLRDarkerBlue] (radius) at (axis cs:350,0.65) 
            {\large\textbf{r}};
            \node[red] (dtheta) at (axis cs:22.5,0.94) {$\Delta \theta$};
            \node[DLRDarkerGray] (dphi) at (axis cs:67.5,0.9) {$\Delta 
            \varphi$};
        \end{polaraxis}
    \end{tikzpicture}
    \caption{Offset strategy for the optimality test with track point indices in 
    blue.}
    \label{fig:opt_test}
\end{figure}
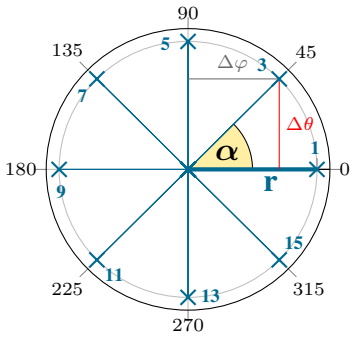

\begin{figure}
\centering
\begin{tikzpicture}
\begin{axis}[%
    width=\columnwidth,
    height=.5\columnwidth,
    restrict x to domain=33300:37800,
    scaled x ticks=false,
    xtick={0, 1800, 3600, 5400, 7200, 9000, 10800, 12600, 
        14400, 16200, 18000, 19800, 21600, 23400, 25200, 27000, 
        28800, 30600, 32400, 34200, 36000, 37800, 39600, 41400, 
        43200, 45000, 46800, 48600, 50400, 52200, 54000, 55800, 
        57600, 59400, 61200, 63000, 64800, 66600, 68400, 70200, 
        72000, 73800, 75600, 77400, 79200, 81000, 82800, 84600},
    xticklabels={{00:00}, {00:30}, {01:00}, {01:30}, {02:00}, 
        {02:30}, {03:00}, {03:30}, {04:00}, {04:30}, {05:00}, 
        {05:30}, {06:00}, {06:30}, {07:00}, {07:30}, {08:00}, 
        {08:30}, {09:00}, {09:30}, {10:00}, {10:30}, {11:00}, 
        {11:30}, {12:00}, {12:30}, {13:00}, {13:30}, {14:00}, 
        {14:30}, {15:00}, {15:30}, {16:00}, {16:30}, {17:00}, 
        {17:30}, {18:00}, {18:30}, {19:00}, {19:30}, {20:00}, 
        {20:30}, {21:00}, {21:30}, {22:00}, {22:30}, {23:00}, 
        {23:30}},
    x tick label style={rotate=90},
    xlabel={UTC [HH:MM]},
    ylabel={Signal Level [dBm]},
    xmin=33300,
    xmax=37800,
    ymax=-32,
    ymin=-39,
    grid=both,
    minor tick num=1,
    /tikz/font=\tiny,
    legend columns=3,
    legend style={/tikz/every even column/.append style={column 
            sep=1em},
        font=\scriptsize},
    legend image post style={line width=1mm},
    legend cell align={left},
    legend to name=opt_cycle_legend]
\addlegendimage{no markers, DLRDarkGrey, ultra thick}
\addlegendentry{Signal Level}
\addlegendimage{no markers, red!70!black, ultra thick}
\addlegendentry{Smoothed Signal Level}
\addlegendimage{no markers, DLRDarkerBlue, ultra thick, dashed}
\addlegendentry{Learned Pointing}
%
\draw [DLRDarkerBlue, dashed, thick] ({axis cs:33391,-40}) -- ({axis 
    cs:33391,-32});
\draw [DLRDarkerBlue, dashed, thick] ({axis cs:33921,-40}) -- ({axis 
    cs:33921,-32});
\draw [DLRDarkerBlue, dashed, thick] ({axis cs:34451,-40}) -- ({axis 
    cs:34451,-32});
\draw [DLRDarkerBlue, dashed, thick] ({axis cs:34981,-40}) -- ({axis 
    cs:34981,-32});
\draw [DLRDarkerBlue, dashed, thick] ({axis cs:35511,-40}) -- ({axis 
    cs:35511,-32});
\draw [DLRDarkerBlue, dashed, thick] ({axis cs:36041,-40}) -- ({axis 
    cs:36041,-32});
\draw [DLRDarkerBlue, dashed, thick] ({axis cs:36571,-40}) -- ({axis 
    cs:36571,-32});
\draw [DLRDarkerBlue, dashed, thick] ({axis cs:37101,-40}) -- ({axis 
    cs:37101,-32});
\draw [DLRDarkerBlue, dashed, thick] ({axis cs:37631,-40}) -- ({axis 
    cs:37631,-32});
\addplot+[DLRDarkGrey, mark=none, thick] table [x={time}, y={if_lvl}]
{dscovr_2403008.tab};
\addplot+[red!70!black, mark=none] table [x={time}, y={if_lvl}]
{dscovr_2403008_smooth_25.0.tab};
\end{axis}
\end{tikzpicture}
\vskip 1ex
\ref{opt_cycle_legend}
\caption{Measured signal level for one offset cycle in our optimality test.}
\label{fig:lvl_cycle_opt_test}
\end{figure}
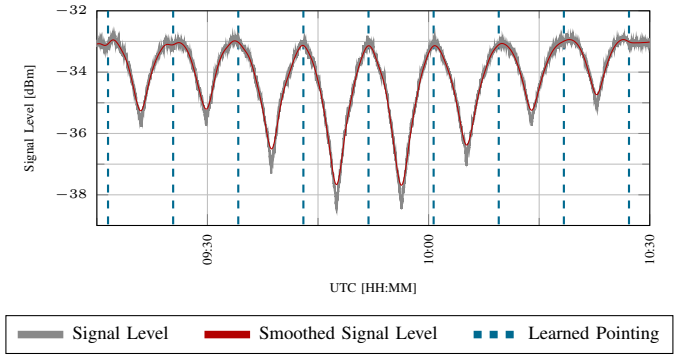

Subsequently, we demonstrate that the learned antenna pointing correction is 
optimal and works well in practice.
For doing so, we create another tracking table for operational data reception 
of the DSCOVR satellite. It implements the learned pointing directions as well 
as an offset strategy which we apply on top.
We show the idea of setting azimuth and elevation angle offsets in 
Fig.~\ref{fig:opt_test}.
The center denotes the previously learned antenna pointing direction.
In our test, we deviate from the latter in a cyclic manner.
Our cycle consists of 17 sequential track points. Starting with index 0 every 
second point implements the learned pointing direction (no offset).
In all points with uneven index, the pointing directions deviate by
$r = 0.75^\circ$. This value is chosen in accordance with the beamwidth of the 
antenna to prevent a loss of signal.
The offset direction -- in terms of the azimuth-elevation plane -- changes by 
$\alpha 
= 45^\circ$ between the track points which implement an offset.
In this way, we finish a complete offset cycle after 17 track points.
As before, the antenna implements a linear interpolation between all tracking 
points.
One can see the corresponding signal level measurements for one such test cycle in 
Fig.~\ref{fig:lvl_cycle_opt_test}.
It is important to note, that the highest signal level occurs very close to the 
learned antenna pointing (i.e. for track points with no offset).
In terms of the local maximum signal level, the mean absolute error for the 
learned pointing directions is $0.085124$~dBm, the maximum error is 
$0.360258$~dBm. We estimate these errors using a minimally smoothed signal 
level (employing a Gaussian filter with $\sigma = 5$ for smoothing).   
From our point of view, these errors are negligible, especially due to noise 
and environmental effects like obstacles on the signal level 
measurements.


\section{Summary and Outlook}
\label{sec:sum}
In our paper, we have shown an efficient method for antenna pointing 
recalibration in an operational environment.
The proposed calibration strategy implements a robust selection process for 
training data which is used to estimate a suitable coordinate transform.
Within our context, a setup involving a linear transformation is appropriate.
In principle, our approach can easily be adapted to 
support different coordinate transformations.
From a practical point of view, the usage of signal level measurements -- which 
are a byproduct of the operational satellite data reception -- is a welcome 
feature. It reduces the need for additional hardware and work steps for antenna 
pointing calibration.
In our future work, we investigate the application of our method to other 
antenna systems. Amongst others, this requires the consideration of a higher 
number of degrees 
of freedom. Furthermore, we want to evaluate the necessity of more 
comprehensive coordinate transformations w.r.t. different possible sources of 
antenna pointing errors.
%


\section*{Acknowledgment}

I thank my colleagues at the German Remote Sensing Data Center for valuable 
comments, fruitful discussions, and technical assistance.


\bibliographystyle{IEEEtran}
\bibliography{IEEEabrv,eusipco_2024.bib}

\end{document}